\documentclass[10pt,twocolumn,letterpaper]{article}

\usepackage[pagenumbers]{wacv} 

\definecolor{wacvblue}{rgb}{0.21,0.49,0.74}
\usepackage[pagebackref,breaklinks,colorlinks,allcolors=wacvblue]{hyperref}
\usepackage{multirow}

\def\wacvPaperID{3015} 
\def\confName{WACV}
\def\confYear{2027}

\title{Dyna3: VLM-Guided Training-Free 4D Reconstruction via Depth Foundation Models}

\author{Xinhao Xiang$^{1}$, Weiyang Li$^{2}$, Zhijie Zheng$^{1}$, Abhijeet Rastogi$^{1}$, Jiawei Zhang$^{1}$\\
$^{1}$IFM Lab, University of California, Davis, CA, USA\\
$^{2}$Genies Inc., CA, USA\\
{\tt\small \{xhxiang, jiwzhang\}@ucdavis.edu, wli@genies.com}
}

\begin{document}
\maketitle
\begin{abstract}

Recent depth foundation models like Depth Anything 3 (DA3) achieve remarkable multi-view depth estimation but assume static 3D scenes, limiting their applicability to real-world dynamic environments. Existing training-free 4D methods like Easi3R and VGGT4D rely on correspondence-trained backbones whose attention encodes cross-frame matching, a property absent in depth-only models like DA3. We present Dyna3, a training-free framework that extends DA3 for 4D dynamic scene reconstruction without any fine-tuning. Our key insight is that DA3's cross-view features, though trained only for depth consistency, implicitly encode motion-discriminative signals when combined with best-match feature search across frames. Its static surfaces find consistent matches globally, while dynamic objects cannot. We further adopt vision-language models (VLM) to automatically generate scene-specific semantic prompts for SAM 3, enabling precise instance-level segmentation that distinguishes which objects move from what objects exist. For reconstruction, we decouple the scene into a cross-frame aligned static background and per-frame dynamic point clouds. 
Experiments on four datasets demonstrate that 
Dyna3 surpasses correspondence-trained methods with +5.5pp J-Mean over 
state-of-the-art VGGT4D on dynamic object segmentation, while achieving up to 13× faster pose estimation 
and 3× faster 4D reconstruction with 4 to 8× lower memory. Dyna3 could therefore enable 
much denser temporal sampling that prior methods cannot support.

\end{abstract}
    
\section{Introduction}
\label{sec:intro}


\begin{figure*}[t]
  \centering
  \includegraphics[width=\linewidth]{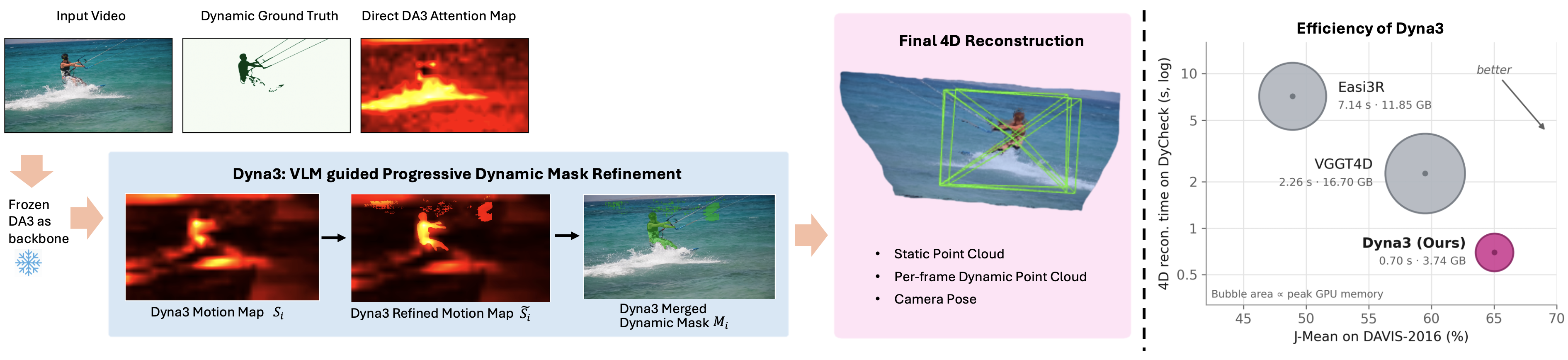}
  \caption{Dyna3 extends a frozen depth foundation model (DA3) to training-free 4D reconstruction.
  \textit{Left:} DA3's attention misses moving objects. Dyna3 uses VLM-guided progressive mask refinement to reconstruct static scene, per-frame dynamic points, and camera poses.
  \textit{Right:} Dyna3 is more accurate than Easi3R and VGGT4D, $3.2\times$ faster and $4.5\times$ less memory than VGGT4D.}
  \label{fig:teasor}
\end{figure*}

Reconstructing dynamic 3D scenes from monocular video is a fundamental challenge in computer vision with applications in autonomous driving, augmented reality, and robotics. Real-world scenes are inherently dynamic: people walk, vehicles move, and objects deform. However, most 3D reconstruction pipelines assume a rigid, static world. Bridging this gap without expensive per-scene optimization or motion supervision remains an open problem.

Recent depth foundation models have advanced 3D reconstruction significantly. Depth Anything 3 (DA3)~\cite{depth_anything_3}, built on a frozen DINOv2~\cite{dinov2} backbone with dual DPT heads, jointly predicts depth, camera rays, and poses in a single feed-forward pass. Compared to correspondence-based models such as DUSt3R~\cite{dust3r} and VGGT~\cite{vggt}, DA3 uses a smaller architecture (135M vs 571M--1.3B parameters) and avoids iterative optimization, enabling efficient inference suitable for real-time applications~\cite{EffiPerception}.

Extending 3D reconstruction to dynamic scenes without from-scratch training typically follows two approaches. The first mines motion cues from attention patterns in correspondence-trained backbones. For example, Easi3R~\cite{easi3r} extracts dynamic-vs-static signals from DUSt3R's cross-attention maps, while VGGT4D~\cite{vggt4d} analyzes VGGT's attention across multiple frames. These methods are training-free but inherit the computational cost of their underlying backbones and fundamentally rely on correspondence-supervised attention that encodes cross-frame matching. The second approach, exemplified by MonST3R~\cite{monst3r}, fine-tunes DUSt3R on dynamic data with explicit tracking objectives ($\mathcal{L}_{\text{track}}$), introducing training cost and dataset dependency.

DA3 fits neither approach. It is trained with geometric objectives ($\mathcal{L}_{\text{depth}}, \mathcal{L}_{\text{ray}}, \mathcal{L}_{\text{pcd}}, \mathcal{L}_{\text{pose}}$) for depth consistency rather than cross-frame correspondence. Its attention maps encode depth relationships, not matching patterns, so directly applying attention-based motion detection yields only ${\sim}8\%$ J-Mean (shown in Sec.~\ref{sec:ablation}). When applied to dynamic scenes without adaptation, moving objects are treated as static geometry, corrupting both scene structure and camera trajectory.  \textbf{Could it be extended to training-free 4D dynamic scene reconstruction while preserving its efficiency advantages?}

We demonstrate that it can. Rather than analyzing attention patterns, we identify that DA3's patch features, though trained for depth, still encode appearance information useful for matching. We introduce a \emph{best-match feature search} that asks whether each patch can find a similar appearance anywhere in neighboring frames. Static surfaces find consistent matches across frames regardless of camera motion, while dynamic objects exhibit low match scores because their motion disrupts cross-frame consistency. However, feature-based signals alone conflate motion with texture variance, revealing the need for semantic guidance.

To address this, we introduce Dyna3, a training-free framework that couples DA3's geometry with instance-level reasoning (Fig.~\ref{fig:teasor}). Rather than relying on manually defined category priors, Dyna3 uses a vision-language model (VLM) to automatically generate scene-specific text prompts, which drive SAM3\cite{sam3} to produce precise instance masks of candidate dynamic objects. This semantic guidance identifies \emph{what} objects exist, while our motion signal determines \emph{which} ones move. For reconstruction, Dyna3 decouples the scene into two components: (1) a cross-frame aligned static background, accumulated from every frame's static pixels via DA3 poses and voxel-consolidated into a single persistent model, and (2) per-frame dynamic point clouds that animate over time. The entire pipeline requires \emph{no} modification to DA3's weights or architecture.

Experiments on four dynamic datasets demonstrate superior performance in 
dynamic object segmentation, camera pose estimation, and 4D reconstruction. 
Dyna3 achieves $65.04\%$ J-Mean on DAVIS-2016~\cite{davis}, surpassing state-of-the-art 
VGGT4D by $+5.5$ percentage points. On pose estimation, Dyna3 achieves 
competitive accuracy while running up to $13\times$ faster with $8\times$ 
less memory. 
On final 4D reconstruction, Dyna3 also reaches comparable quality with 
$3\times$ faster inference and $4\times$ less memory. These efficiency gains 
allow Dyna3 to process much denser temporal sampling where prior methods could run out 
of memory.
Our contributions are summarized as follows:
\begin{itemize}

    
    \item We present Dyna3, the first training-free framework that extends a depth-only foundation model (DA3) to 4D dynamic scene reconstruction while preserving its efficiency advantage over correspondence-based methods.

    \item We design a unified pipeline that introduce best-match feature search for motion detection, VLM-guided SAM~3 segmentation for instance separation, and decoupled static and dynamic point clouds for coherent reconstruction.

    \item We achieve state-of-the-art training-free results on dynamic object segmentation with significant efficiency gains on camera pose estimation and 4D reconstruction, enabling much denser temporal sampling than prior methods support.

\end{itemize}

\section{Related Work}
\label{sec:related}


\subsection{Feed-Forward Multi-View 3D Reconstruction}

Traditional multi-view reconstruction methods such as COLMAP~\cite{colmap} rely on iterative optimization for camera calibration and dense matching, limiting real-time applicability. DUSt3R~\cite{dust3r} introduced a feed-forward paradigm by directly regressing aligned point maps from image pairs. MASt3R~\cite{mast3r} extended it with dense feature matching, and subsequent work scaled this paradigm to many views and streaming settings, including Fast3R~\cite{fast3r}, Spann3R~\cite{spann3r}, CUT3R~\cite{cut3r}, and TTSA3R~\cite{TTSA3R}. VGGT~\cite{vggt} generalized the pointmap formulation into a geometry-grounded transformer predicting depth, cameras, and tracks jointly, adding an explicit tracking loss $\mathcal{L}_{\text{track}}$ for correspondence supervision. These methods achieve accurate geometry but use large models (571M--1.3B parameters) and often require iterative optimization. Depth Anything3 (DA3)\cite{depth_anything_3} takes a different path, unifying multi-view depth, camera rays, and pose into a single feed-forward model built on a frozen DINOv2~\cite{dinov2} backbone with just about 135M parameters. DA3 is trained only for depth consistency without correspondence supervision, enabling faster inference but leaving open whether it can support dynamic reconstruction. While feed-forward geometric reconstruction has also advanced in single-image settings~\cite{PRISM}, extending to dynamic multi-view scenes remains challenging.

\subsection{Dynamic and 4D Scene Reconstruction}

While 3D perception has advanced through multi-modal fusion for tasks like object detection~\cite{TransFusion, BEVFusion, FusionViT, 3DifFusionDet, OVODA}, dynamic scene reconstruction presents distinct challenges.
Classical dynamic reconstruction relies on per-scene optimization, from deformable neural radiance fields~\cite{dnerf,nerfies} to dynamic Gaussian splatting~\cite{deformable_gs,f4dgs} and monocular 4D methods such as Shape of Motion~\cite{shapeofmotion}. These achieve high fidelity but require lengthy optimization and often known poses. A complementary line adapts feed-forward pointmap models to motion. MonST3R~\cite{monst3r} fine-tunes DUSt3R on dynamic data, introducing training cost and dataset dependency. Easi3R~\cite{easi3r} is training-free, repurposing DUSt3R's cross-attention maps to distinguish dynamic pixels, but is restricted to two-view inputs and inherits DUSt3R's computational overhead. VGGT4D~\cite{vggt4d} is another training-free method mining motion cues from VGGT's attention for multi-view 4D reconstruction with stronger accuracy, but remains expensive. In addition, these training-free methods rely on correspondence-trained backbones whose attention encodes matching. On the other hand, depth-only models like DA3 offer faster inference, but lack such attention-based motion cues. Our work addresses this gap by extracting motion signals through best-match feature search and leveraging vision-language models for instance-level segmentation, enabling efficient training-free 4D reconstruction without correspondence-trained backbones.

\section{Method}
\label{sec:method}

\begin{figure*}[t]
  \centering
  \includegraphics[width=\linewidth]{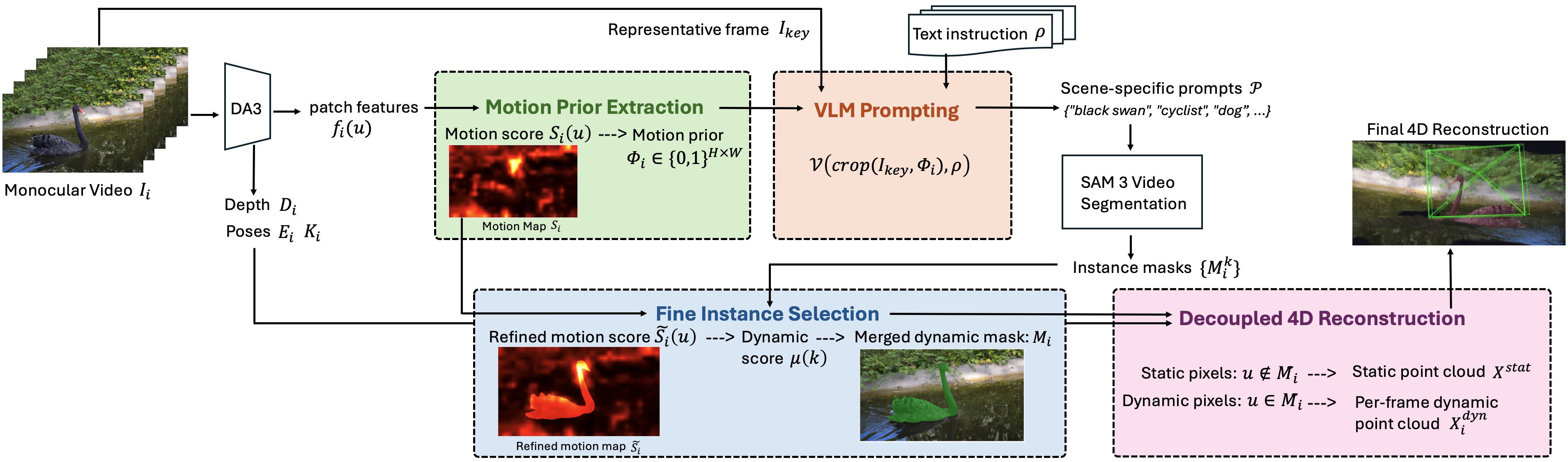}
  \caption{Overview of the Dyna3.}
  \label{fig:dyna3}
\end{figure*}

\subsection{Overview}
Given a monocular video $\{I_i\}_{i=1}^{N}$ of a dynamic scene, Dyna3 produces a 4D reconstruction consisting of a persistent static background, per-frame dynamic objects, and camera poses. Our framework operates entirely at inference time using a pretrained DA3 model, with no fine-tuning or additional training. Shown in Fig.\ref{fig:dyna3}, the pipeline has four stages. First, we extract per-frame depth, camera pose, and patch features from DA3\cite{depth_anything_3} (Sec.\ref{sec:prelim}). 
Second, we compute motion score $S_i$ from patch similarities via best-match search, producing a coarse motion prior $\Phi_i$ that identifies regions containing movement (Sec.\ref{sec:motion}). Third, we use this motion prior to guide a vision-language model, which generates scene-specific text prompts for SAM3\cite{sam3} to produce precise instance masks $M_i$ (Sec.\ref{sec:vlm}).
Fourth, we fuse the motion score signal with the instance masks to select dynamic objects (Sec.\ref{sec:select}) and decouple the scene into static and dynamic components for 4D reconstruction (Sec.\ref{sec:recon}).

\subsection{Geometry Extraction from DA3}
\label{sec:prelim}

We adopt DA3~\cite{depth_anything_3} as the geometry backbone. For a set of input views, DA3 predicts per-frame depth $D_i \in \mathbb{R}^{H\times W}$ and a dense ray map that encodes per-pixel camera rays. 
DA3 provides two options for camera pose estimation: a camera-token head and RANSAC~\cite{ransac} decomposition of the ray map. We use the latter, which decomposes the ray map into extrinsics $E_i \in SE(3)$ and intrinsics $K_i$, as we find it more reliable across diverse scenes (see Sec.~\ref{sec:exp}).

DA3's DPT decoder also produces feature maps $F_i \in \mathbb{R}^{h \times w \times C}$, where $f_i(u)$ denotes the feature vector at spatial location $u$. These features, originally designed for depth regression, encode rich appearance information that we repurpose for motion detection as described in Sec.~\ref{sec:motion}.

Given depth $D_i$ and camera parameters $(K_i, E_i)$, each pixel $u$ is back-projected to a world coordinate:
\begin{equation}
X_i(u) \;=\; E_i \,\cdot\, \pi^{-1}\!\big(u,\, D_i(u);\, K_i\big),
\label{eq:unproject}
\end{equation}
where $\pi^{-1}$ denotes standard pinhole unprojection. This lifts every pixel into a unified world frame for scene reconstruction.

\subsection{Motion Feature via Best-Match Search}
\label{sec:motion}

Existing training-free 4D methods such as Easi3R~\cite{easi3r} and VGGT4D~\cite{vggt4d} detect motion by analyzing attention patterns from correspondence-trained backbones. In these models, attention weights concentrate on matching regions for static content but scatter for dynamic objects that violate cross-frame consistency. Easi3R aggregates mean and variance statistics from DUSt3R's cross-attention maps, while VGGT4D computes similarity matrices from VGGT's query-key representations. Both approaches succeed because correspondence-supervised training produces attention distributions that inherently distinguish static from dynamic content.

DA3's attention maps do not carry such motion-discriminative signals. DA3 is trained for depth consistency rather than cross-frame correspondence, so its global self-attention encodes depth relationships instead of matching patterns (see Sec.~\ref{sec:ablation} for analysis). However, we found that DA3's patch features still encode appearance information useful for matching. We exploit this by constructing a motion signal through explicit feature comparison rather than attention analysis.

For each location $u$ in frame $i$, we search for its best-matching location across a temporal neighborhood $\mathcal{N}_i$ by computing cosine similarity between feature vectors. A static surface appears consistently across frames and produces high similarity to its best match, even when camera motion shifts its image location. A dynamic object, however, changes appearance or position in ways that break this consistency, resulting in lower match scores. We define the motion score at location $u$ in frame $i$ as:
\begin{equation}
  S_i(u) = 1 - \frac{1}{|\mathcal{N}_i|} \sum_{j\in\mathcal{N}_i} \max_{u' \in \Omega_j} \frac{\langle f_i(u), f_j(u') \rangle}{\|f_i(u)\| \|f_j(u')\|}
  \label{eq:motion}
\end{equation}
where $\mathcal{N}_i$ is the temporal neighborhood of frame $i$, $\Omega_j$ is the set of all spatial locations in frame $j$, and the maximization finds the highest cosine similarity across all locations in each neighbor frame.
This global search naturally handles camera motion, as a static patch will match its displaced counterpart regardless of where it appears in neighboring frames. A high score $S_i(u)$ indicates that location $u$ cannot find good matches and is likely dynamic.
We normalize $S_i$ per frame to handle scene-dependent similarity distributions, threshold at $\tau$, and keep the largest connected component to suppress isolated noise. This produces a coarse motion prior $\Phi_i \in \{0,1\}^{H\times W}$ that identifies regions containing moving content. 

The motion prior is robust across diverse scenes, though as an appearance-based signal it provides spatially coarse localization rather than precise object boundaries. We use it as the first step to identify which objects move, and then adopt segmentation for accurate boundary delineation.

\subsection{VLM-Guided Instance Segmentation}
\label{sec:vlm}

The motion prior $\Phi_i$ from Sec.~\ref{sec:motion} identifies regions containing dynamic content but does not provide object-level boundaries. A straightforward fix approach would apply fixed category labels such as ``person'' or ``vehicle'' to segment potential dynamic objects, but this fails when moving objects span unpredictable categories or when generic labels match both moving and stationary instances. We instead use the motion prior to guide a vision-language model (VLM): by focusing the VLM on regions where motion is detected, it can identify scene-specific object descriptions rather than relying on predefined categories. Such VLM-guided video understanding has shown effectiveness in various vision tasks~\cite{BVQI, MaxVQA, RQ_VQA, ADGVE, AIGVE_Tool, LMM_VQA}.

We extract a bounding box from the motion prior $\Phi_i$ and crop the corresponding region from a representative frame $I_{\text{key}}$. We then query the VLM with this cropped region and a text instruction $\rho$ that asks the model to identify the main moving subject (see supplementary for prompt details). The instruction is designed to generate specific, descriptive noun phrases rather than broad category names. The VLM returns a set of scene-specific prompts:
\begin{equation}
  \mathcal{P} \;=\; \mathcal{V}\big(\text{crop}(I_{\text{key}}, \Phi_i),\, \rho\big).
\end{equation}
For example, on a street scene the model might return \texttt{\{"cyclist","dog"\}} rather than generic \texttt{"person"} or \texttt{"animal"}. This scene-adaptive prompting produces more discriminative queries that better isolate the actual moving entities.

Each prompt $c \in \mathcal{P}$ is then passed to SAM3~\cite{sam3}, a video segmentation model that tracks instances across frames. For each frame $i$, 
SAM3 returns instance masks $M_i^{c,k}$, where $c$ is the prompt and $k$ indexes instances matching that prompt. Each mask $M_i^{c,k} \in\{0,1\}^{H \times W}$ indicates the region of instance $k$.
Original SAM3 computes video features independently of the text prompt, which is inefficient. We cache the video representation once per clip and reuse the them for each subsequent prompt, clearing only the prompt-conditioned state. This yields a $4\times$ speedup over per-prompt re-encoding while producing equivalent results.
We collect all instance masks across prompts into a candidate set $\{M_i^k\}$ for motion-aware selection.

\subsection{Motion-Aware Dynamic Instance Selection}
\label{sec:select}

The VLM-guided segmentation produces candidate instance masks, but not all candidates correspond to moving objects. Semantic categories often include both dynamic and static instances. For example, a scene with multiple people may contain one walking figure and several stationary bystanders. Selecting all ``person'' instances would incorrectly include the static ones in the dynamic reconstruction.

We resolve this ambiguity by combining the instance
masks $M^k_i$ with the motion score $S_i(u)$: the motion score
indicates how likely each pixel is to be dynamic, while the
instance masks provide precise object boundaries. 
We define the refined motion score map $\tilde{S}_i \in \mathbb{R}^{H \times W}$ by masking $S_i$ to instance regions. 
For each
candidate instance $k$ with mask $M^k_i$, we compute a dynamic score by averaging the motion values within
the mask:

\begin{equation}
  \mu(k) = \operatorname*{mean}_{u \in M_i^k} S_i(u)
\end{equation}
Instances with $\mu(k)$ above threshold $\theta$ are classified as dynamic. When all scores fall below $\theta$, we select the instance with the highest score to ensure at least one dynamic object is identified.

The selected instances are merged to form the per-frame dynamic mask:
\begin{equation}
  M_i = \bigcup_{k:\mu(k)>\theta} M_i^k
\end{equation}
This approach effectively combines coarse motion localization with precise instance boundaries, producing accurate dynamic masks for 4D reconstruction.

\subsection{Decoupled 4D Scene Reconstruction}
\label{sec:recon}

With per-frame dynamic masks $M_i$ established, the final 4D representation is constructed by lifting classified pixels into 3D and organizing them into static and dynamic components. 
Separating static and dynamic content ensures the background remains clean without moving objects, while dynamic content is separately preserved with temporal correspondence.

\paragraph{Static background.}
The static background aggregates observations from all frames, excluding regions marked as dynamic.
At dynamic object boundaries, segmentation masks may not perfectly align with the true dynamic object extent, causing pixels to leak into the static model and create ghost artifacts. 
To handle such imperfect mask boundaries, each dynamic mask is dilated by $\delta$ pixels to create a safety margin: $\tilde{M}_i = \mathrm{dilate}(M_i, \delta)$. 
Static frame pixels (those outside $\tilde{M}_i$) are then projected into 3D coordinates using Eq.~\ref{eq:unproject} and aggregate them into a unified point cloud. Since multiple frames observe the same static surfaces from different viewpoints, this aggregation produces many redundant points at identical 3D locations. Voxel-grid consolidation is applied to merge these redundant observations:
\begin{equation}
  \mathcal{X}^{\text{stat}} \;=\;
  \operatorname{Voxel}_{v}\!\Big(
  \textstyle\bigcup_{i=1}^{N}\,
  \big\{\, X_i(u) \;:\; u \notin \tilde{M}_i \,\big\}\Big).
  \label{eq:static}
\end{equation}
where $\operatorname{Voxel}_{v}$ denotes voxel-grid downsampling~\cite{open3d} at resolution $v$, keeping one representative point per voxel. This reduces point cloud density while preserving geometric detail. Aggregating observations from multiple frames also improves coverage, as each viewpoint reveals surfaces occluded from other positions~\cite{OccFace}.

\paragraph{Dynamic foreground.}
Unlike the static background, dynamic objects cannot be merged across frames because their positions change over time. Each frame's dynamic content is therefore maintained as a separate point cloud:
\begin{equation}
  \mathcal{X}^{\text{dyn}}_i \;=\;
  \big\{\, X_i(u) \;:\; u \in M_i \,\big\},
  \qquad i = 1,\dots,N.
  \label{eq:dynamic}
\end{equation}
Each $\mathcal{X}^{\text{dyn}}_i$ captures the 3D geometry of moving objects at their position in frame $i$. Rendering these point clouds in sequence produces the animation of dynamic content against the fixed background.

\paragraph{Camera poses and rendering.}
The camera poses $E_i$ are obtained from DA3's ray map via RANSAC decomposition as described in Sec.~\ref{sec:prelim}. Together with the point clouds, they form the complete 4D reconstruction: the static point cloud $\mathcal{X}^{\text{stat}}$, the sequence of dynamic point clouds $\mathcal{X}^{\text{stat}}$, and the camera poses $E_i$. The camera path is visualized as frustums colored by temporal progression through the sequence. To render the 4D reconstruction, the static point cloud is displayed as a persistent background with each dynamic point cloud $\mathcal{X}^{\text{dyn}}_i$ overlaid at its corresponding timestep $i$.

\section{Experiment}
\label{sec:exp}

We evaluate Dyna3 on three tasks: dynamic object segmentation 
(Section~\ref{sec:seg_results}), camera pose estimation (Section~\ref{sec:pose_results}), 
and 4D reconstruction (Section~\ref{sec:4d_results}).

\subsection{Experimental Setup}
\textbf{Datasets.} We use DAVIS-2016 and DAVIS-2017~\cite{davis} for dynamic object segmentation, 
TUM-dynamics~\cite{tum} and Sintel~\cite{sintel} for camera pose estimation, 
and DyCheck~\cite{dycheck} for 4D reconstruction.
\textbf{Baselines.} We compare against MonST3R~\cite{monst3r}, DAS3R~\cite{das3r}, 
CUT3R~\cite{cut3r}, Easi3R~\cite{easi3r}, and VGGT4D~\cite{vggt4d}. 
MonST3R, DAS3R, and CUT3R are fine-tuned on dynamic datasets with optical 
flow supervision. Easi3R and VGGT4D are training-free like ours, but rely on 
correspondence-trained backbones whose attention encodes cross-frame matching. 
In contrast, Dyna3 extends a depth-only model without such correspondence priors.
\textbf{Metrics.} For dynamic object segmentation, we report J-Mean and F-Mean following DAVIS convention. 
For camera pose estimation, we report Absolute Trajectory Error (ATE), Relative 
Translation Error (RTE), and Relative Rotation Error (RRE). For 4D 
reconstruction, we report Accuracy, Completeness, and Distance. We also 
report inference time and peak memory usage to evaluate efficiency.
\textbf{Implementation.} Dyna3 uses DA3-BASE with a frozen DINOv2 ViT-Base encoder, 
Qwen2-VL-2B~\cite{qwen2vl} as the VLM, and SAM~3~\cite{sam3} for video 
segmentation. Motion scores are computed from last layer features with a 75th 
percentile threshold. Camera poses are derived from ray maps via RANSAC. 
All experiments run on a single NVIDIA A100 GPU.


\subsection{Dynamic Object Segmentation}
\label{sec:seg_results}

Dynamic object segmentation evaluates whether the method can accurately 
identify moving regions, which is essential for separating foreground from 
background in 4D reconstruction.  Table~\ref{tab:davis} 
compares against training-free baselines.

\begin{table}[t]
\centering
\caption{
Dynamic object segmentation on DAVIS-2016 and DAVIS-2017. J-Mean and F-Mean denote region similarity and boundary 
accuracy (\%), respectively. Best results are \textbf{bold}. Dyna3 achieves state-of-the-art performance on 
both datasets, with particularly large gains on boundary accuracy.}
\label{tab:davis}
\resizebox{\linewidth}{!}{
\begin{tabular}{lcccccc}
\toprule
& \multicolumn{2}{c}{DAVIS-16} & \multicolumn{2}{c}{DAVIS-17} \\
\cmidrule(lr){2-3} \cmidrule(lr){4-5}
Method & J-Mean$\uparrow$ & F-Mean$\uparrow$ & J-Mean$\uparrow$ & F-Mean$\uparrow$ \\
\midrule
Easi3R~\cite{easi3r} & 48.92 & 45.29 & 49.65 & 44.09 \\
VGGT4D~\cite{vggt4d} & 59.50 & 55.47 & 56.45 & 51.09 \\
\textbf{Dyna3 (Ours)} & \textbf{65.04} & \textbf{66.93} & \textbf{58.73} & \textbf{62.46} \\
\bottomrule
\end{tabular}
}
\end{table}

Dyna3 achieves 65.04\% J-Mean on DAVIS-2016, outperforming state-of-the-art 
VGGT4D by +5.5 percentage points and Easi3R by +16.1 percentage points. 
The gains are even larger on boundary accuracy, with +11.5pp F-Mean 
improvement over VGGT4D. This suggests that SAM~3's video segmentation 
produces sharper object boundaries than the point-cloud-based masks from 
correspondence methods. On the more challenging DAVIS-2017 with multiple 
objects per video, Dyna3 maintains its lead with +2.3pp J-Mean and +11.4pp 
F-Mean over VGGT4D. These results demonstrate that best-match feature search 
provides a robust motion signal even without correspondence training, and 
that VLM-guided instance selection effectively distinguishes which objects 
move from what objects exist in the scene.

\subsection{Camera Pose Estimation}
\label{sec:pose_results}

Accurate camera pose estimation is critical for 4D reconstruction, as pose 
errors propagate to point cloud alignment. We evaluate on TUM-dynamics, which 
contains indoor scenes with walking people, and Sintel, which features complex 
camera motion with dynamic characters. Tables~\ref{tab:tum} and~\ref{tab:sintel} 
compare pose accuracy and computational cost.

\begin{table}[t]
\centering
\caption{
Camera pose estimation on TUM-dynamics. ATE and RTE are in meters, 
RRE in degrees. Best and second-best results are \textbf{bold} and 
\underline{underlined}. Dyna3 achieves competitive pose accuracy while being 
significantly more efficient, running 2.9$\times$ faster with 5.3$\times$ less 
memory than VGGT4D.}
\label{tab:tum}
\resizebox{\linewidth}{!}{
\begin{tabular}{lccccc}
\toprule
Method & ATE$\downarrow$ & RTE$\downarrow$ & RRE$\downarrow$ & Time(s/vid)$\downarrow$ & Mem(GB)$\downarrow$ \\
\midrule
MonST3R~\cite{monst3r} & 0.156 & 0.103 & 12.041 & 82.93 & 42.4 \\
Easi3R~\cite{easi3r} & \underline{0.063} & 0.046 & 2.523 & 138.27 & 45.7 \\
VGGT4D~\cite{vggt4d} & \textbf{0.016} & \textbf{0.020} & \textbf{0.612} & 45.53 & 130.1 \\
\textbf{Dyna3 (Ours)} & 0.121 & \underline{0.024} & \underline{2.506} & \textbf{15.64} & \textbf{24.7} \\
\bottomrule
\end{tabular}
}
\end{table}

\begin{table}[t]
\centering
\caption{Camera pose estimation on Sintel.
Camera pose estimation on Sintel. ATE and RTE are in meters, RRE in 
degrees. Best and second-best results are \textbf{bold} and \underline{underlined}. 
Dyna3 achieves the best relative translation error while running 8.6$\times$ 
faster and using 4.4$\times$ less memory than VGGT4D.}
\label{tab:sintel}
\resizebox{\linewidth}{!}{
\begin{tabular}{lccccc}
\toprule
Method & ATE$\downarrow$ & RTE$\downarrow$ & RRE$\downarrow$ & Time(s/vid)$\downarrow$ & Mem(GB)$\downarrow$ \\
\midrule
Easi3R~\cite{easi3r} & \underline{0.151} & 0.051 & \underline{0.277} & 220.05 & 27.7 \\
VGGT4D~\cite{vggt4d} & \textbf{0.076} & \underline{0.046} & \textbf{0.273} & 28.03 & 45.2 \\
\textbf{Dyna3 (Ours)} & 0.197 & \textbf{0.040} & 0.319 & \textbf{3.26} & \textbf{10.2} \\
\bottomrule
\end{tabular}
}
\end{table}

On TUM-dynamics, Dyna3 achieves competitive relative pose accuracy, ranking 
second in both RTE and RRE, while significantly outperforming 
all baselines in efficiency. Compared to VGGT4D, Dyna3 runs 2.9$\times$ faster 
(15.6s vs 45.5s) and uses 5.3$\times$ less memory (24.7GB vs 130.1GB). Compared 
to Easi3R, the speedup is even more significant at 8.8$\times$ faster with 
1.8$\times$ less memory. While VGGT4D achieves lower ATE due to its 
correspondence-trained backbone that explicitly optimizes cross-frame 
consistency, Dyna3's DA3-based mechanism inherits the efficiency of depth-only 
models, and our pipeline successfully preserves this advantage while extending 
DA3 to handle dynamic scenes.

On Sintel, Dyna3 achieves the best relative translation error, 
outperforming both VGGT4D and Easi3R. This demonstrates that 
DA3's ray-based pose estimation, combined with our motion-aware static 
background separation, produces accurate frame-to-frame motion even without 
correspondence training. The efficiency gains are also significant: Dyna3 runs 
8.6$\times$ faster than VGGT4D (3.26s vs 28.03s) and 67$\times$ faster than 
Easi3R (3.26s vs 220s), while using 4.4$\times$ and 2.7$\times$ less memory 
respectively. This efficiency allows Dyna3 to process longer videos and denser 
temporal sampling that would be impractical with correspondence-based methods.

\paragraph{Efficiency enables dense temporal sampling.}
The key advantage of Dyna3's efficiency is enabling denser frame sampling that 
correspondence-based methods cannot support due to their quadratic memory 
scaling. Tables~\ref{tab:stride_tum} and~\ref{tab:stride_sintel} show how 
temporal stride affects pose accuracy.

\begin{table*}[t]
\centering
\caption{
Effect of temporal stride on TUM-dynamics. Denser sampling (\textit{i.e.} smaller 
stride) significantly improves relative pose accuracy. At stride 3, VGGT4D 
runs out of memory while Dyna3 completes successfully, demonstrating the 
practical advantage of efficient inference.}
\label{tab:stride_tum}
\resizebox{0.65\linewidth}{!}{
\begin{tabular}{lcccccc}
\toprule
Stride & ATE$\downarrow$ & RTE$\downarrow$ & RRE$\downarrow$ & Time (vs VGGT4D)$\downarrow$ & Mem (vs VGGT4D)$\downarrow$ \\
\midrule
30 & 0.023 & 0.140 & 13.674 & 3.63 (12.4$\times$ faster) & 16.4 (7.9$\times$ smaller)\\
10 & 0.021 & 0.063 & 9.232 & 7.57 (13.4$\times$ faster) & 21.5 (6.7$\times$ smaller)\\
3 & 0.021 & 0.024 & 2.506 & 15.64 (VGGT4D OOM) & 24.7 (VGGT4D OOM)\\
\bottomrule
\end{tabular}
}
\end{table*}

\begin{table*}[t]
\centering
\caption{Effect of temporal stride on Sintel. Denser sampling (\textit{i.e.} smaller 
stride) dramatically 
reduces rotation error (RRE improves 7.5$\times$ from stride 3 to stride 1). 
Dyna3 remains 8-10$\times$ faster than VGGT4D across all stride settings.}
\label{tab:stride_sintel}
\resizebox{0.65\linewidth}{!}{
\begin{tabular}{lcccccc}
\toprule
Stride & ATE$\downarrow$ & RTE$\downarrow$ & RRE$\downarrow$ & Time (vs VGGT4D)$\downarrow$ & Mem (vs VGGT4D)$\downarrow$ \\
\midrule
3 & 0.208 & 0.106 & 2.383 & 1.15 (9.4$\times$ faster) & 6.4 (5.0$\times$ smaller) \\
2 & 0.208 & 0.092 & 1.698 & 1.61 (10.0$\times$ faster) & 8.6 (5.1$\times$ smaller) \\
1 & 0.197 & 0.040 & 0.319 & 3.26 (8.6$\times$ faster) & 10.2 (4.4$\times$ smaller) \\
\bottomrule
\end{tabular}
}
\end{table*}

On TUM-dynamics, reducing temporal sampling stride from 30 to 3 dramatically improves 
relative pose accuracy: RTE decreases by 5.8$\times$ (from 0.140m to 0.024m) 
and RRE decreases by 5.5$\times$ (from 13.7° to 2.5°). This improvement is 
expected since denser sampling provides more constraints for pose estimation 
and reduces the motion between consecutive frames. Crucially, VGGT4D runs out 
of memory at stride 3 due to its quadratic memory scaling with frame count, 
while Dyna3 completes successfully in 15.6 seconds using only 24.7GB memory. 
Even at the sparsest stride-30 setting where both methods can run, Dyna3 
remains 12.4$\times$ faster and 7.9$\times$ more memory-efficient.

On Sintel, the benefits of dense sampling are also much apparent. Reducing 
stride from 3 to 1 improves RRE by 7.5$\times$ (from 2.38° to 0.32°) and RTE 
by 2.7$\times$ (from 0.106m to 0.040m). At stride-1, Dyna3 achieves the best 
RTE among all methods while still running 8.6$\times$ faster and using 
4.4$\times$ less memory than VGGT4D at its sparsest feasible stride. These 
results demonstrate that Dyna3's efficiency advantage directly translates into 
accuracy gains: by enabling temporal sampling densities that correspondence-based 
methods cannot handle, Dyna3 achieves competitive or superior pose accuracy 
while maintaining substantial computational savings.

This efficiency-accuracy trade-off provides users flexibility to adapt Dyna3 
to different application requirements. In latency-sensitive scenarios such as 
real-time robotics or interactive AR/VR applications, users can set a larger 
temporal stride to minimize inference time and memory footprint while still 
obtaining reasonable pose estimates. Conversely, in offline processing 
scenarios such as video editing or high-fidelity 3D content creation where 
accuracy matters the most, users can use denser temporal sampling to maximize 
pose precision. This flexibility is unique to Dyna3's efficient architecture 
and is not available to correspondence-based methods, which are constrained 
to sparse sampling due to their memory and inference time limitations.

\subsection{4D Reconstruction}
\label{sec:4d_results}

The ultimate goal of our pipeline is to produce high-quality 4D point cloud 
reconstructions that accurately capture both static background and dynamic 
foreground. We evaluate on DyCheck~\cite{dycheck}, which contains iPhone-captured 
dynamic scenes with ground truth point clouds from multi-view capture. 
Table~\ref{tab:dycheck} compares reconstruction quality and computational cost.


\begin{table*}[t]
\centering
\caption{
4D reconstruction on DyCheck. Accuracy measures distance from 
predicted to ground truth points; Completeness measures the reverse direction; 
Distance combines both. Best and second-best results are \textbf{bold} and 
\underline{underlined}. Dyna3 achieves the second-best reconstruction quality 
while being 3.2$\times$ faster and using 4.5$\times$ less memory than VGGT4D, 
and outperforms all fine-tuned methods (MonST3R, CUT3R, DAS3R) despite being 
training-free.
}
\label{tab:dycheck}
\resizebox{0.9\linewidth}{!}{
\begin{tabular}{l|cc|cc|cc|cc}
\toprule
\multirow{2}{*}{\textbf{Method}} & \multicolumn{2}{c|}{\textbf{Accuracy}$\downarrow$} & \multicolumn{2}{c|}{\textbf{Completeness}$\downarrow$} & \multicolumn{2}{c|}{\textbf{Distance}$\downarrow$} & \multirow{2}{*}{\textbf{Time}$\downarrow$} & \multirow{2}{*}{\textbf{Mem.}$\downarrow$} \\
& Mean & Med. & Mean & Med. & Mean & Med. & & \\
\midrule
DAS3R~\cite{das3r} & 0.192 & 0.142 & 0.250 & 0.108 & 0.428 & 0.336 & - & - \\
CUT3R~\cite{cut3r} & 0.073 & 0.054 & 0.133 & 0.049 & 0.328 & 0.224 & - & - \\
MonST3R~\cite{monst3r} & 0.090 & \underline{0.033} & 0.113 & 0.064 & 0.279 & 0.234 & - & - \\
Easi3R~\cite{easi3r} & 0.070 & 0.044 & 0.060 & 0.033 & 0.194 & \underline{0.132}  & 7.14s (10.2$\times$ slower) & 11.85GB (3.2$\times$ larger)\\
VGGT4D~\cite{vggt4d} & \textbf{0.022} & \textbf{0.004} & \textbf{0.051} & \textbf{0.012} & \textbf{0.123} & \textbf{0.050} & 2.26s (3.2$\times$ slower) & 16.70GB (4.5$\times$ larger) \\
\midrule
\textbf{Dyna3 (Ours)} & \underline{0.065} & 0.038 & \underline{0.058} & \underline{0.028} & \underline{0.180} & 0.135 & \textbf{0.70s} & \textbf{3.74GB} \\
\bottomrule
\end{tabular}
}
\end{table*}

Dyna3 achieves the second-best reconstruction quality across most metrics while 
being substantially more efficient than all baselines. On Accuracy, Dyna3 
achieves 0.065 mean error, ranking second only to VGGT4D and 
outperforming all fine-tuned methods including MonST3R, CUT3R, 
and DAS3R. On Completeness, which measures coverage of the ground truth 
surface, Dyna3 achieves 0.058 mean error, again ranking second and notably 
outperforming MonST3R and CUT3R by a large margin. This 
demonstrates that our training-free approach, despite using a depth-only 
backbone without correspondence supervision, produces more complete 
reconstructions than methods specifically fine-tuned on dynamic datasets.

The efficiency gains on 4D reconstruction are still significant. Dyna3 processes each 
video in 0.70 seconds, which is 3.2$\times$ faster than VGGT4D and 
10.2$\times$ faster than Easi3R. Memory usage is even more favorable: 
Dyna3 requires only 3.74GB, compared to 16.7GB for VGGT4D (4.5$\times$ larger) 
and 11.85GB for Easi3R (3.2$\times$ larger). This low memory footprint enables 
Dyna3 to run on consumer-grade GPUs and process high-resolution videos that 
would exceed memory limits on correspondence-based methods.

The quality gap between Dyna3 and VGGT4D primarily reflects the difference in 
how the two methods handle dynamic regions. VGGT4D uses its correspondence-trained 
backbone to track points across frames, enabling temporally consistent dynamic 
reconstruction. In contrast, Dyna3's decoupled representation reconstructs 
dynamic objects per-frame without cross-frame correspondence, which limits 
temporal coherence but avoids error accumulation from incorrect matches. 
Importantly, our decoupled static and dynamic separation ensures that reconstruction 
errors in dynamic regions do not propagate to the static background, which makes up the majority of most scenes. This design choice trades some dynamic 
reconstruction quality for robustness and efficiency, enabling Dyna3 to deliver 
competitive overall quality at a fraction of the computational cost.

\subsection{Ablation Studies}
\label{sec:ablation}


\begin{table}[t]
\centering
\caption{Ablation study on Dyna3 components on DAVIS-2016. We progressively add components 
to validate their contribution. Each component provides consistent improvements 
in dynamic object segmentation.}
\label{tab:ablation}
\resizebox{\linewidth}{!}{
\begin{tabular}{lcc}
\toprule
\textbf{Variant} & J-Mean$\uparrow$ & F-Mean$\uparrow$ \\
\midrule
DA3 baseline (no motion pipeline) & 8.16 & 8.43 \\
+ Motion Prior Extraction & 29.65 & 27.34 \\
+ Fine Instance Selection & 33.51 & 32.18 \\
+ VLM Prompting (w/o SAM3) & 53.35 & 54.29 \\
+ SAM3 Segmentation (\textbf{Full Dyna3}) & \textbf{65.04} & \textbf{66.93} \\
\bottomrule
\end{tabular}
}
\end{table}

We validate the contribution of each component in Dyna3 by \textit{progressively} 
adding them and measuring segmentation performance on DAVIS-2016. 
Table~\ref{tab:ablation} summarizes the results.
\textbf{DA3 baseline.} We run DA3 for depth estimation and treat all pixels 
as static, producing a uniform zero mask with no dynamic region detection. 
This achieves only 8.16\% J-Mean, confirming that DA3's depth-trained 
features do not directly encode motion information without explicit 
comparison across frames.
\textbf{+ Motion Prior Extraction.} We add our best-match feature search 
to compute motion scores $S_i(u)$, threshold at the 75th percentile to 
obtain the coarse motion prior $\Phi_i$, and use $\Phi_i$ directly as the 
final dynamic mask. This improves J-Mean to 29.65\% (+21.5pp), validating 
that DA3's patch features encode appearance information useful for motion 
detection through explicit cross-frame comparison.
\textbf{+ Fine Instance Selection.} We then refine the coarse motion 
prior by extracting connected components and selecting the largest one as 
the dynamic region. This provides a modest improvement to 33.51\% (+3.9pp), 
but fails when multiple objects exist or when the moving object is not the 
largest region in the scene.
\textbf{+ VLM Prompting (w/o SAM3).} We next crop the key frame using the motion 
prior bounding box, feed it to the VLM to generate scene-specific prompts, 
and use the VLM's attention map or bounding box output directly as the 
dynamic mask without SAM3 refinement. This significantly improves performance 
to 53.35\% (+19.8pp), demonstrating that feature-based motion signals alone 
could confuse motion with texture variance, and semantic guidance from VLMs is 
necessary to identify what objects are actually moving.
\textbf{+ SAM3 Segmentation (Full Dyna3).} We finally pass VLM-generated prompts to 
SAM3 for precise video segmentation, then apply our motion-aware instance 
selection using $\mu(k) = \text{mean}_{u \in M_i^k} S_i(u)$ to select which 
instances are actually moving. This achieves 65.04\% J-Mean (+11.7pp) and 
66.93\% F-Mean. The large F-Mean improvement confirms that SAM3 provides 
significantly sharper boundaries than VLM-based masks alone.

\subsection{Analysis: How DA3's attention layers does not encode motion feature}
\label{sec:attention_analysis}

\begin{figure}[t]
\centering
\includegraphics[width=\linewidth]{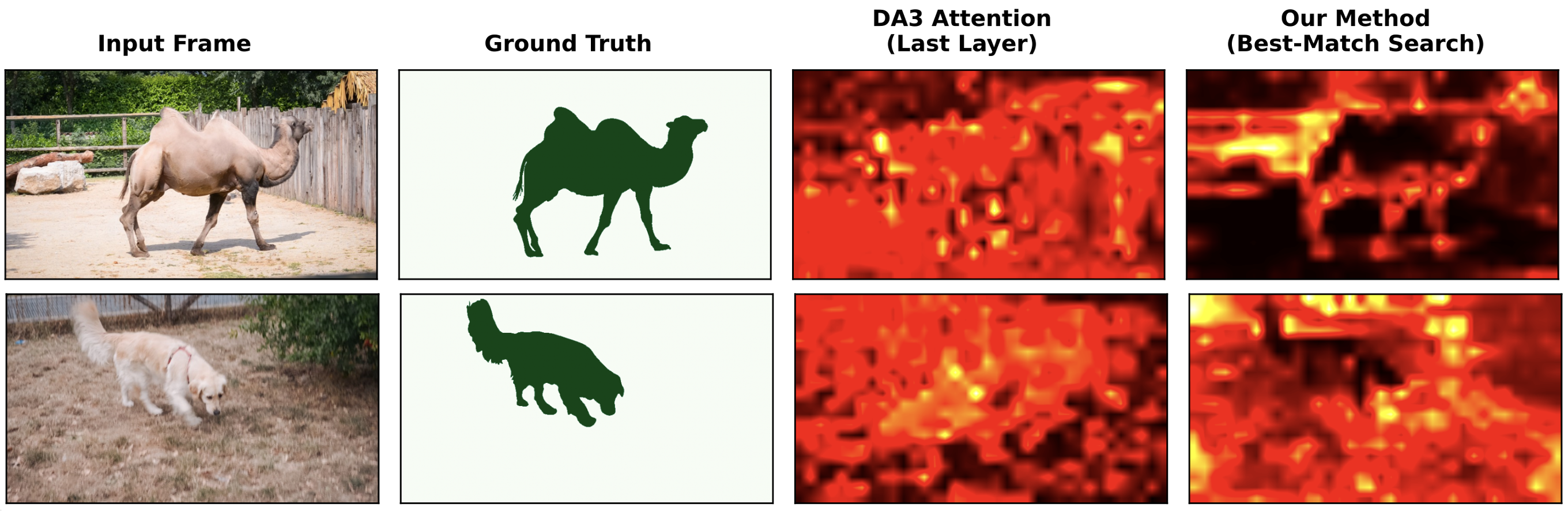}
\caption{Comparison of DA3 cross-attention maps versus our best-match feature 
search on DAVIS-2016. DA3's attention (Column 3) produces uniformly distributed 
responses across the scene, failing to isolate dynamic objects (Column 2). 
In contrast, our method (Column 4) successfully concentrates on moving regions. 
This pattern holds consistently across all 20 validation videos.}
\label{fig:attention_vs_motion}
\end{figure}

In Section~\ref{sec:method}, we claimed that DA3's attention does not carry 
motion-discriminative signals, which is why we use explicit feature comparison 
instead of attention-based detection as baselines~\cite{vggt4d, easi3r}. Figure~\ref{fig:attention_vs_motion} 
provides qualitative evidence for this claim.
We extract cross-attention maps from DA3's last layer and compute attention 
statistics following Easi3R's formulation. For correspondence-trained models 
like VGGT and DUSt3R, this approach successfully highlights dynamic regions 
because their attention learns cross-frame matching patterns during training. 
However, DA3 was trained with depth objectives ($\mathcal{L}_{\text{depth}}$, 
$\mathcal{L}_{\text{ray}}$, $\mathcal{L}_{\text{pcd}}$, $\mathcal{L}_{\text{pose}}$) 
that optimize for geometric consistency rather than correspondence. As shown 
in Figure~\ref{fig:attention_vs_motion}, DA3's attention maps are uniformly 
distributed across the scene and do not correlate with ground truth dynamic 
masks.

This motivates our design choice: rather than analyzing attention patterns, 
we construct motion signals through explicit feature comparison. Our best-match 
feature search asks whether each patch can find a similar appearance anywhere 
in neighboring frames. Static surfaces find consistent matches regardless of 
camera motion, while dynamic objects exhibit low match scores. This approach
achieves 29.65\% J-Mean using only the motion prior, compared to attention-based 
detection which achieves only ${\sim}8\%$, which is barely above random.

\section{Conclusion}
\label{sec:conclusion}

We presented Dyna3, a training-free framework that extends Depth Anything 3 
to 4D dynamic scene reconstruction. Our key contributions are best-match 
feature search for motion detection in depth-trained models and VLM-guided 
instance segmentation for accurate dynamic object identification. Dyna3 
achieves 65.04\% J-Mean on DAVIS-2016, outperforming state-of-the-art VGGT4D 
by +5.5 percentage points while running up to 13$\times$ faster with 8$\times$ 
less memory. This efficiency enables dense temporal sampling that prior methods 
cannot support, allowing Dyna3 to achieve competitive pose accuracy where 
correspondence-based methods run out of memory.

{
    \small
    \bibliographystyle{ieeenat_fullname}
    \bibliography{main}
}
\maketitlesupplementary
\setcounter{page}{1}
\setcounter{table}{8}
\setcounter{figure}{4}
\setcounter{section}{0}
\renewcommand\thesection{\Alph{section}}

\section{VLM Prompt Details}
\label{sec:vlm_prompt}
We use Qwen2-VL-7B~\cite{qwen2vl} as the vision-language model. Given the motion prior $\Phi_i$, we extract a bounding box and crop the corresponding region from the representative frame $I_{\text{key}}$, the middle frame of the sequence. We query the VLM with the following prompt:
\begin{quote}
\texttt{In this video frame, it contains something that is moving on the static scene. Please identify the main moving foreground subject in it. Answer with 1-3 specific noun phrases. Do not use generic terms like "person" or "object".
}
\end{quote}
The VLM output is directly used as the text prompt for SAM 3 video segmentation.


\section{Progressive Dynamic Mask Refinement}
\label{sec:progressive}

\begin{figure*}[t]
\centering
\includegraphics[width=\textwidth]{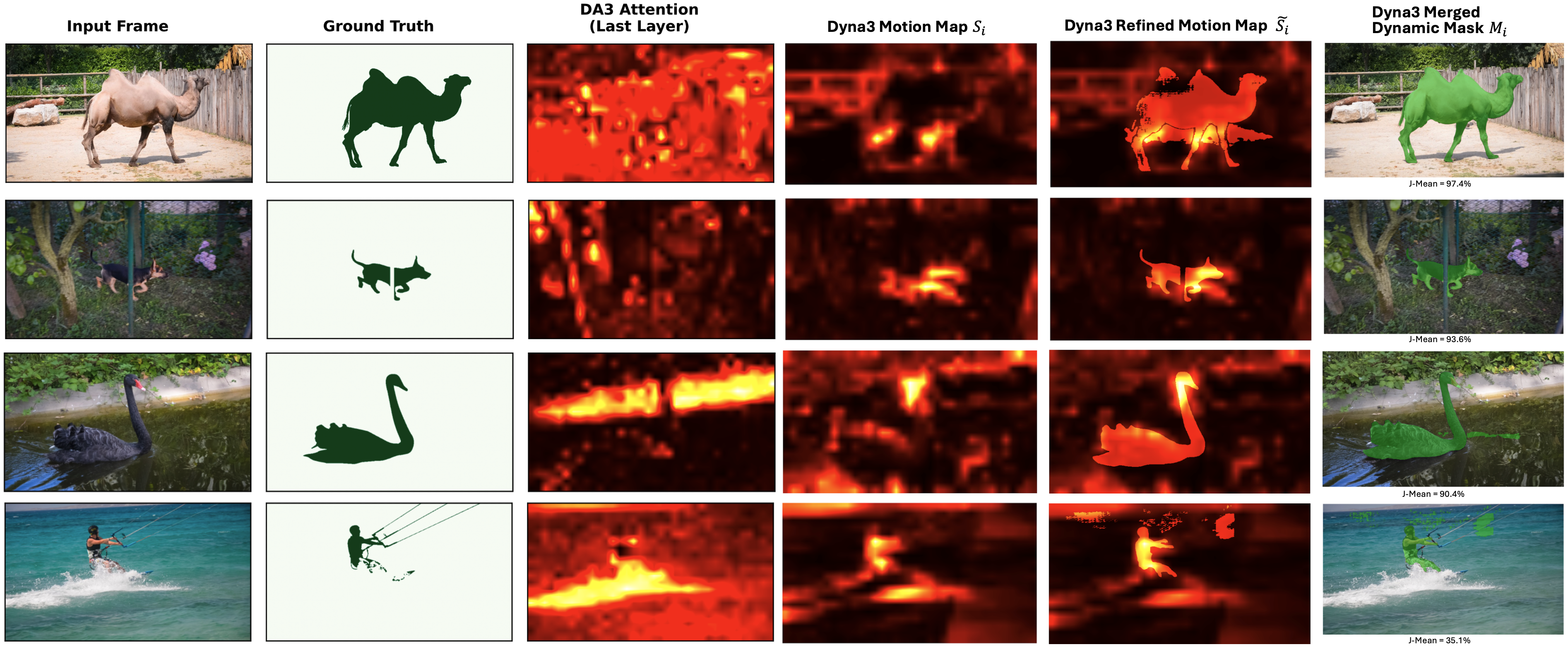}
\caption{Progressive dynamic mask refinement on DAVIS-2016. Column 1-2: input frame and ground truth. Column 3: DA3 cross-attention map computed following Easi3R's formulation, which fails to isolate dynamic objects due to DA3's depth-focused training. Columns 4-6: Dyna3's progressive refinement from motion map $S_i$ (best-match feature search), to refined motion map $\tilde{S}_i$ (masked by SAM3 instance regions), to final merged dynamic mask $M_i$ (after motion-aware instance selection). Per-video J-Mean scores demonstrate strong performance on clear subjects while revealing challenges on small objects.}
\label{fig:progressive}
\end{figure*}

Figure~\ref{fig:progressive} illustrates how Dyna3 progressively refines dynamic object detection through each pipeline stage. We provide detailed explanations of how each visualization is generated and analyze how the results connect to our method architecture.

\noindent\textbf{DA3 Attention Map.}
Following Easi3R's formulation~\cite{easi3r}, we extract cross-attention maps from DA3's last global layer. For each pixel in the target frame, we compute attention statistics over reference frames: $\mu$ (maximum attention value indicating match confidence) and $\sigma$ (attention spread indicating uncertainty). The dynamic score is computed as $(1-\mu) \cdot \sigma$, where low match confidence and high spread suggest dynamic content. However, as shown in Column 3, DA3's attention produces uniformly distributed responses across the scene. This failure is a result of DA3's training objective: DA3~\cite{depth_anything_3} is trained with depth consistency losses ($\mathcal{L}_{\text{depth}}$, $\mathcal{L}_{\text{ray}}$, $\mathcal{L}_{\text{pcd}}$, $\mathcal{L}_{\text{pose}}$) rather than correspondence supervision ($\mathcal{L}_{\text{track}}$), so its attention encodes depth relationships instead of cross-frame matching patterns.

\noindent\textbf{Motion Map $S_i$.}
Instead of directly analyzing attention, we construct motion signals through explicit feature comparison. For each pixel $u$ in frame $i$, we search for its best-matching location across temporal neighbors using cosine similarity between DA3's patch features (Eq. 2 in main paper). 
The motion map $S_i$ aggregates these scores across all pixels. Static surfaces find consistent matches regardless of camera motion, producing low scores, while dynamic objects exhibit high scores due to appearance changes across frames. As shown in Column 4, $S_i$ successfully highlights dynamic regions but includes background noise from textured areas.

\noindent\textbf{Refined Motion Map $\tilde{S}_i$.}
The refined motion map $\tilde{S}_i$ masks $S_i$ to SAM3 instance regions, combining our motion signals with semantic boundaries. SAM3 provides precise instance masks $\{M_i^k\}$ guided by VLM-generated prompts, and $\tilde{S}_i$ restricts the motion signal to these semantic regions. As shown in Column 5, this refinement eliminates background noise while preserving motion information within object boundaries. The improvement from $S_i$ to $\tilde{S}_i$ demonstrates the value of semantic guidance: motion detection alone may confuse texture variance with actual movement, while instance-level reasoning could isolates true dynamic objects.

\noindent\textbf{Merged Dynamic Mask $M_i$.}
The final mask $M_i$ is produced by motion-aware instance selection. For each instance $k$, we compute the dynamic score $\mu(k) = \text{mean}_{u \in M_i^k} S_i(u)$ and select instances exceeding threshold $\theta$. As shown in Column 6, $M_i$ provides clean binary segmentation that closely matches ground truth. This stage resolves ambiguity when SAM3 segments multiple instances of the same category. Only instances with high motion scores are included, effectively distinguishing moving objects from static ones.\\

\noindent The progressive improvement from $S_i$ to $\tilde{S}_i$ to $M_i$ validates our pipeline design. Each component addresses a specific limitation: best-match feature search overcomes DA3's lack of correspondence-trained attention, VLM-guided SAM3 segmentation provides semantic boundaries that motion signals alone cannot produce, and motion-aware instance selection resolves category-level ambiguity. Our method achieves strong results on sequences with clear, distinct subjects (camel 97.4\%, dog 93.6\%, blackswan 90.4\%), while still facing challenges on small or distant objects (kite-surf 35.1\%) where both motion detection and instance segmentation become less reliable.

\section{Why Attention-Based Motion Detection Fails on DA3}
\label{sec:attention_analysis_sup}

Existing training-free 4D methods extract motion-discriminative signals from attention patterns in correspondence-trained backbones. This section analyzes the architectural and training differences that explain why such approaches succeed on DUSt3R~\cite{dust3r} and VGGT~\cite{vggt} but fail on DA3~\cite{depth_anything_3}x.

\paragraph{Correspondence-Trained Attention.}
DUSt3R use explicit cross-attention between source and reference views with separate query, key, and value projections:
\begin{equation}
A_{i \to j} = \text{softmax}\left(\frac{Q_i K_j^\top}{\sqrt{d}}\right)
\end{equation}
where $Q_i$ derives from view $i$ and $K_j$ from view $j$. Under correspondence supervision ($\mathcal{L}_{\text{track}}$), the network learns to concentrate attention on matching locations for static content. Dynamic objects violate this learned correspondence, causing attention to scatter. Easi3R~\cite{easi3r} exploits this property by computing per-pixel statistics:
\begin{equation}
A_{\text{dyn}} = (1 - \mu_{\text{src}}) \cdot \sigma_{\text{src}} \cdot \mu_{\text{ref}} \cdot (1 - \sigma_{\text{ref}})
\end{equation}
where $\mu$ denotes maximum attention (match confidence) and $\sigma$ denotes attention spread (uncertainty). Low confidence combined with high spread indicates correspondence failure, signaling dynamic content.

VGGT extends this principle to multi-view settings. Its transformer aggregator processes all views jointly with explicit tracking supervision, producing attention distributions that encode cross-frame correspondences. VGGT4D~\cite{vggt4d} analyzes these attention patterns to identify regions where learned correspondences break down.

\paragraph{Depth-Trained Attention.}
DA3 employs a fundamentally different architecture. Its global self-attention concatenates tokens from all views into a single sequence:
\begin{equation}
A = \text{softmax}\left(\frac{Q \cdot [K_1; K_2; \ldots; K_N]^\top}{\sqrt{d}}\right)
\end{equation}
where $[\cdot;\cdot]$ denotes concatenation. Unlike cross-attention, this formulation does not separate query and key by view, and more critically, DA3 lacks correspondence supervision entirely.

\paragraph{Training Objective Comparison.}
Table~\ref{tab:training_loss} summarizes the training losses across architectures. The presence or absence of $\mathcal{L}_{\text{track}}$ determines whether attention encodes correspondence patterns.

\begin{table}[h]
\centering
\caption{Training loss comparison across architectures.}
\label{tab:training_loss}
\begin{tabular}{lccc}
\toprule
Loss Component & DUSt3R & VGGT & DA3 \\
\midrule
$\mathcal{L}_{\text{depth}}$ (depth supervision) & \checkmark & \checkmark & \checkmark \\
$\mathcal{L}_{\text{camera}}$ (pose supervision) & \checkmark & \checkmark & \checkmark \\
$\mathcal{L}_{\text{pmap}}$ (direct pointmap) & \checkmark & \checkmark & -- \\
$\mathcal{L}_{\text{pcd}}$ (point cloud) & -- & -- & \checkmark \\
$\mathcal{L}_{\text{track}}$ (correspondence) & \checkmark & \checkmark & -- \\
\bottomrule
\end{tabular}
\end{table}

DUSt3R and VGGT include $\mathcal{L}_{\text{track}}$, which supervises 2D point correspondences across frames. This loss forces the attention mechanism to learn cross-frame matching. Given a query location, the network must attend to its corresponding location in other views. Consequently, attention patterns directly encode correspondence quality. Strong, focused attention indicates reliable matches, while scattered attention indicates correspondence failure.

DA3 optimizes for depth consistency through $\mathcal{L}_{\text{depth}}$, $\mathcal{L}_{\text{ray}}$, $\mathcal{L}_{\text{pcd}}$, and $\mathcal{L}_{\text{pose}}$. These objectives encourage geometric coherence across views but impose no constraint on where attention should focus. The resulting attention patterns encode depth relationships and scene structure rather than point-to-point correspondences.

\paragraph{Implication for Motion Detection.}
The absence of correspondence supervision in DA3 renders attention-based motion detection ineffective. Applying Easi3R's formulation to DA3's attention produces uniformly distributed responses unrelated to dynamic content, as demonstrated in Figure~\ref{fig:progressive}. This motivates our approach: rather than analyzing attention patterns, we construct motion signals through explicit feature comparison via best-match search, which does not rely on correspondence-trained representations.

\section{Additional Implementation Details}
\label{sec:impl_details}


Beyond the settings described in Section 4.1, we use temporal neighborhood size $|\mathcal{N}_i| = 4$ frames for motion score computation, dynamic score threshold $\theta = 0.3$ for instance selection, and processing resolution of 504 pixels on the longer side. We exploit SAM3's session reuse to cache video features across multiple text prompts, yielding 4$\times$ speedup over per-prompt re-encoding.

\end{document}